\pdfoutput=1
\documentclass[11pt]{article}
\usepackage{pifont}

\usepackage[]{EMNLP2023}

\usepackage{times}
\usepackage{amsmath}
\usepackage{latexsym}
\usepackage{tabularx}
\usepackage{multirow}
\usepackage{booktabs}
\usepackage{amssymb}
\usepackage[most]{tcolorbox}

\usepackage[T1]{fontenc}
\usepackage[utf8]{inputenc}

\usepackage{microtype}
\usepackage{array}
\usepackage{multirow}
\usepackage{graphicx}
\usepackage{xcolor}
\usepackage{colortbl}
\usepackage{arydshln}

\usepackage{inconsolata}

\title{When Long Helps Short: How Context Length in Supervised Fine-tuning Affects Behavior of Large Language Models}

\author{
Yingming Zheng\textsuperscript{1,3}\thanks{~~Equal contribution.} \quad
Hanqi Li\textsuperscript{1,3,4}\footnotemark[1] \quad
Kai Yu\textsuperscript{1,3,4} \quad
Lu Chen\textsuperscript{1,2,3,4}\thanks{~~Corresponding author.} \\
\textsuperscript{1}X-LANCE Lab, MoE Key Lab of Artificial Intelligence, AI Institute, \\
School of Computer Science, Shanghai Jiao Tong University, Shanghai, China \\
\textsuperscript{2}Shanghai Innovation Institution, Shanghai, China \\
\textsuperscript{3}Jiangsu Key Lab of Language Computing, Suzhou, China \\
\textsuperscript{4}Suzhou Laboratory, Suzhou, China \\
\texttt{\{zhengyingming, daqige, chenlusz\}@sjtu.edu.cn}
}

\begin{document}
\maketitle
\begin{abstract}
Large language models (LLMs) have achieved impressive performance across natural language processing (NLP) tasks. As real-world applications increasingly demand longer context windows, continued pretraining and supervised fine-tuning (SFT) on long-context data has become a common approach. While the effects of data length in continued pretraining have been extensively studied, their implications for SFT remain unclear. In this work, we systematically investigate how SFT data length influences LLM behavior on short-context tasks. Counterintuitively, we find that long-context SFT improves short-context performance, contrary to the commonly observed degradation from long-context pretraining. To uncover the underlying mechanisms of this phenomenon, we first decouple and analyze two key components, Multi-Head Attention (MHA) and Feed-Forward Network (FFN), and show that both independently benefit from long-context SFT. We further study their interaction and reveal a knowledge preference bias: long-context SFT promotes contextual knowledge, while short-context SFT favors parametric knowledge, making exclusive reliance on long-context SFT suboptimal. Finally, we demonstrate that hybrid training mitigates this bias, offering explainable guidance for fine-tuning LLMs.

\end{abstract}
\section{Introduction}
Large Language Models have demonstrated remarkable potential in NLP and achieved substantial success across real-world applications~\cite{NEURIPS2020_1457c0d6,openai2024gpt4technicalreport,zheng2025craveconflictingreasoningapproach}. As application scenarios diversify and task complexity intensifies, there is an increasing demand for extended context windows in LLMs. Conventionally, enhancing context length capabilities primarily involves two critical phases: continual pretraining and SFT, where models are explicitly adapted to process long-context inputs.

Long-context continual pretraining~\cite{fuyao_24_data} has garnered extensive investigation and established consensus conclusions. Specifically, continued pretraining with long-context data has been shown to consistently impair LLMs' capabilities on short-context tasks. To mitigate this performance degradation, current methodologies advocate incorporating a proportion of short-context data into the pretraining corpus, thereby preserving model performance across short-context applications.

In contrast, previous studies on SFT have not established such consensus conclusions. Numerous works have focused on constructing long-context SFT datasets~\cite{zhao2024longskyworktrainingrecipeefficiently,xiong2025from} to achieve promising performance improvements on long-context tasks. However, whether aligning training data length with evaluation benchmark, essential in continual pretraining, remains equally critical for SFT is still unclear. Two fundamental questions remain unresolved: (1) Does long-context SFT enhance or degrade performance on short-context tasks compared to conventional short-context SFT? (2) What underlying mechanisms drive these potential performance variations across context lengths?

To address these research gaps, we extend the framework established by \citet{prolong-8b} by performing SFT with varied datasets on the same long-context-pretrained base model (Llama-3-8B-ProLong-512k-base), followed by comprehensive evaluation across diverse short-context domain benchmarks. Contrary to conventional expectations, our experiments yield counterintuitive results:
\textbf{Models fine-tuned with long-context data demonstrate measurable improvements in aggregate metrics.}

We proceed to conduct research to explain the underlying mechanisms behind this counterintuitive phenomenon. First, we decouple the two main modules of Transformer-based LLMs, Multi-Head Attention (MHA) and Feed-Forward Network (FFN), to independently test their performance. For the MHA module, we test its capabilities from three perspectives: \textit{module replacement}, \textit{retrieval score analysis}, and \textit{attention entropy analysis}. Our findings reveal that SFT with long-context data enhances the standalone capabilities of MHA across all dimensions. Similarly for the FFN module, we employ \textit{module replacement} and conduct detailed analysis of \textit{FFN activation statistics}, reaching the same conclusion that long-context SFT improved the independent performance of FFN across various metrics. \textbf{Employing long-context data for SFT benefits both MHA and FFN modules}.

To more comprehensively explain the underlying mechanisms and better guide practical applications, we further analyze the interactions between the MHA and FFN modules using a knowledge conflict framework. Our findings reveal deeper insights: \textbf{The length of SFT training data introduces biases in the model’s knowledge preferences}. Specifically, models fine-tuned with long-context SFT data exhibit overconfidence in contextual knowledge, while those trained with short-context SFT data show excessive reliance on parametric knowledge. Although long-context SFT enhances the standalone performance of both MHA and FFN modules, the resulting knowledge preference bias makes exclusive reliance on long-context SFT suboptimal. By adjusting the ratio of long-context to short-context data in SFT training, we find that: \textbf{Hybrid training helps mitigate this bias, thereby enabling potentially superior performance.}

In summary, our contributions are as follows:
\begin{itemize}
    \item We present the first systematic study on how SFT data length impacts short-context reasoning performance. By comparing models fine-tuned on datasets with different text lengths, we show that incorporating long-context data leads to overall performance improvements on a variety of short-context tasks.
    \item We provide an in-depth modular analysis revealing how long-context SFT alters LLM internals. By decoupling and separately analyzing MHA and FFN, we demonstrate that long-context SFT improves the standalone performance of both modules, offering new interpretability insights into how different SFT regimes shape model functionality.
    \item We employ a knowledge-conflict framework to study the interactions of MHA and FFN, revealing that varying SFT data lengths induce knowledge preference bias in models. Further, we demonstrate that hybrid training mitigates these biases, providing explainable guidance for SFT in LLM.
\end{itemize}

\section{Related Work}
\subsection{SFT in Long-context LLMs}
SFT is a critical step in aligning LLMs with human intent, typically using curated instruction-response pairs. In the early development of long-context models, such as Longformer~\cite{longformer}, BigBird~\cite{bigbird} and LongLLaMA~\cite{longllama}, researchers fine-tune models on long-context data to match the extended context window, such as multi-document QA~\cite{10.1162/tacl_a_00023}, scientific articles~\cite{jin2019pubmedqadatasetbiomedicalresearch}, or long dialogue transcripts~\cite{zhong2021qmsumnewbenchmarkquerybased}. These works aimed to improve the model's ability to retain and utilize extended context during inference. \par 
Recent work has shown that fine-tuning long-context LLMs on short-context datasets can be sufficient to achieve strong performance even on long-context tasks~\cite{liu-etal-2024-e2}. For example, \citet{prolong-8b} fine-tuned Llama-3 using the short-context UltraChat dataset to obtain the ProLong-8B model, and reported that incorporating long synthetic instructions did not yield additional gains in long-context scenarios. However, this line of work neither provides further analysis nor evaluates the model's performance on short-context tasks, leaving open the question of whether long-context SFT offers broader benefits.
% In this paper, we revisit this assumption and systematically investigate whether training on long-context instructions can improve short-context reasoning and why.
\subsection{Modular Interpretability of Attention and FFN in Language Model}
Understanding how different components of LLMs contribute to reasoning and knowledge encoding has become a central focus in interpretability research~\cite{goldowskydill2023localizingmodelbehaviorpath,zhang2022optopenpretrainedtransformer,hao2021self}. In Transformer-based architectures, attention and FFN modules are believed to play complementary roles: attention layers primarily handle information routing and selection, while FFNs apply nonlinear transformations that can encode domain or factual knowledge~\cite{internalworksoftransformer,jin-etal-2024-cutting}. \citet{geva} first demonstrated that FFN layers can function as key-value memories, retrieving token-level associations to support factual recall. In addition, \citet{dai-etal-2022-knowledge} identified "knowledge neurons" within FFNs whose activation states are strongly correlated with specific knowledge output, suggesting that FFNs serve as important carriers of parametric knowledge. On the attention side, \citet{elhage2021mathematical} and \citet{wang2024differentiationspecializationattentionheads} revealed that attention heads exhibit functional specialization, such as copying, moving or induction, and play crucial roles in reasoning and task-specific representation structuring. \par 
To better understand how these modules behave under different training conditions, recent work has proposed modular analysis techniques such as layer-level ablation, head attribution, and module replacement. \citet{10.5555/3600270.3601532} developed a method to locate and edit factual knowledge in pretrained LLMs by replacing FFN modules and observing prediction changes. Similarly, \citet{yao2025knowledgecircuitspretrainedtransformers} traced how specific knowledge emerges and propagates through Transformer layers. Although most of these studies focus on models in the pretraining stage, few have examined how SFT, especially with varying instruction lengths, modifies the internal functionality of these modules. In this work, we extend the modular interpretability analysis to the SFT stage, examining how instruction length affects the functionality of attention and FFN modules.

\section{Approach}
\subsection{Preliminary}
In this work, we mainly focus on the autoregressive Transformer-based language models. Given a Transformer decoder with $L$ layers and $N$ heads and an input sequence $x = \{x_1,\dots,x_N\}$ consisting with $N$ tokens, the output of the $l$-th layer can be donated as $H_l = \{h_{l.1},\dots,h_{l.T}\}$. Each Transformer layer consists of a multi-head attention (MHA) module and a FFN module with residual connections connecting them, as shown by the following formula:
\begin{equation}
    \widetilde{\mathbf{H}}_l = \text{MHA}(\mathbf{H}_{l-1}) + \mathbf{H}_{l-1},
\end{equation}
\begin{equation}
    \mathbf{H}_l = \text{FFN}(\widetilde{\mathbf{H}}_l) +\widetilde{\mathbf{H}}_l .
\end{equation}

A MHA block consists of $M$ attention heads, which are capable of aggregating global information into hidden states. In the $i$-th head of the MHA module in the $l$-th layer, the hidden states are first projected into query $\mathbf{Q}_l^i$, key $\mathbf{K}_l^i$ and value $\mathbf{V}_l^i$ matrices. Positional information is then incorporated into the query and key matrices through the RoPE with rotation matrix $\mathbf{R}_\theta$ ($\theta$ is the RoPE base). These matrices are subsequently processed through a dot product followed by a softmax operation to compute the attention scores $A_l^i$. Finally, the value representations are weighted by the attention scores, and all attention heads are concatenated and projected to produce the attention output of $l$-th layer as follows:
\begin{equation}
    A^i_l = \text{Softmax}(\mathbf{Q}_l^{i\mathbf{T}} \mathbf{R}_{\theta} \mathbf{K}_l^i / \sqrt{d}),
\end{equation}
\begin{equation}
    \text{MHA}(\mathbf{H}_{l-1}) = \text{Concat}(\{\mathbf{A}_l^i \mathbf{V}_l^i\}^N_{i=1})\mathbf{W}^O ,
\end{equation}
where d is the dimension of query and key, $\mathbf{W}^O$ is the projection matrix, and $\text{Concat}$ is the concatenation of hidden states.

The FFN in Transformer block is composed of two learnable weight matrices: $\mathbf{W}_{l}^{in}$ and $\mathbf{W}_{l}^{out}$. $\mathbf{W}_{l}^{in}$ reads the residual stream state. Its results passed through element-wise non-linear activation function $g(\cdot)$, producing neuron activations. These are then transformed by $\mathbf{W}_{l}^{out}$ to produce outputs:
\begin{equation}\label{eq:5}
    \text{FFN}(\widetilde{\textbf{H}}_{l}) = g(\widetilde{\textbf{H}}_l \textbf{W}_{l}^{in})\textbf{W}_{l}^{out} .
\end{equation}

\subsection{Module Replacement}
Module replacement is a commonly used method in interpretability analysis, aiming to evaluate the contribution of specific components within a model. Rather than simply removing a module to observe performance degradation, researchers often replace it with an alternative implementation to gain deeper insight into its functional importance and potential for improvement. Moreover, by comparing performance metrics before and after replacement, one can quantitatively assess the relative effectiveness of modules in different models.

\subsection{Retrieval Score}
Followed by the work of \citet{wu2024retrievalheadmechanisticallyexplains}, we calculate the retrieval score of attention heads, which measures its ability to accurately extract relevant information from the context and use it to answer questions during generation. Specifically, given a question-answering instances where the correct answer must be explicitly retrieved from the input passage such as Needle-in-a-Haystack, and we track the overlap between the generated tokens of each head and the source text. Formally, we define:
\begin{equation}
    \text{Retrieval score for head h} = \frac{|g_h \cap k|}{|k|},
\end{equation}
where $g_h$ denotes the set of tokens retrieved by a given attention head $h$, which means the token receives the most attention probability mass by $h$ head as well as the currently generated token of model. The set $k$ represents the correct answer tokens. A higher retrieval score indicates that the attention head is more actively redirecting relevant information from the input.

% \subsection{Attention Entropy}
% Attention head entropy quantifies the sharpness or dispersion of the attention distribution produced by each head~\cite{zhang2024attentionentropykeyfactor}. It is formally defined as:
% \begin{equation}
%     S_i = -\sum_{j=1}^{n} \alpha_{ij}\text{log}\alpha_{ij}
% \end{equation}
% where $\alpha_{ij}$ denotes the attention weight assigned by the $i$-th head to $j$-th position, satisfying $\sum_{j}\alpha_{ij}=1$ and $S_i$ represents the attention Entropy of head $i$. A higher entropy indicates a more uniform distribution over positions, suggesting lower selectivity, while a lower entropy reflects a concentrated focus on a few positions, implying stronger structural sensitivity or specialization.

% \subsection{FFN Activation Statistics}
% FFN Activation denotes $g(\widetilde{\textbf{H}}_l \textbf{W}_l^{in})$ in equation~\ref{eq:5}. Its statistics characterize the magnitude and distribution of activations produced by the FFN layers, offering insights into neuron sparsity and selectivity. Commonly used statistical metrics include the mean, variance, and sparsity of the activation values. The mean reflects the overall activation level, the variance captures the dispersion of activations across neurons, and the sparsity indicates how many neurons are effectively inactive. Together, these metrics help reveal how information is processed and filtered through FFN layers, and whether the model exhibits distributed or localized activation behavior.

\begin{table*}[ht!]
\centering

\resizebox{\textwidth}{!}{
\begin{tabular}{l|ccc|cc|cc|cc|c} 
\hline

\hline
 \multirow{2}{*}{\textbf{SFT Dataset}}&\multicolumn{3}{c|}{\textbf{General}} & \multicolumn{2}{c|}{\textbf{Math}} &  \multicolumn{2}{c|}{\textbf{Code}} & \multicolumn{2}{c|}{\textbf{Commonsense QA}} & \multirow{2}{*}{\textbf{\textit{Avg.}}}                                   \\ 
\cdashline{2-10}[1pt/1pt]
                     & \textit{MMLU}  & \textit{BBH}  & \textit{LAMBADA}  & \textit{GSM8K} & \textit{MATH} &   \textit{MBPP}    & \textit{HUMANEVAL} & \textit{OBQA} & \textit{PIQA}  &
                  \\
\hline
\rowcolor{blue!10} \multicolumn{11}{c}{\emph{Short-Context SFT}}\\
\text{UltraChat} & $61.50$ &	$61.00$&	$64.44$&	$54.69$	&$17.28$	&$39.20$&	$32.92$	&$\textbf{74.40}$	&$75.24$ & $53.41$
\\
\text{Tulu-v2-sft-mixture} & $61.37$  &	$61.50$&	$64.20$&	$62.50$&	$16.42$&$\textbf{51.20}$	&$49.39$&	$\textbf{74.40}$&	$76.33$&	$57.48$\\

\rowcolor{yellow!10} \multicolumn{11}{c}{\emph{Long-Context SFT}}\\

 \text{LongAlpaca} & $62.11$&	$61.29$&	$68.84$&	$\textbf{68.75}$&	$16.70$&	$46.20$&$48.78$	&	$\textbf{74.40}$&	$\textbf{76.77}$ & $58.20$ \\
 \text{LongMIT} & $62.65$&	$61.80$&	$\textbf{71.51}$&	$59.38$&	$\textbf{18.22}$&	$50.20$&	$\textbf{54.27}$&	$74.20$&	$68.28$ & $57.83$ \\
  \text{ChatQA2} & $\textbf{62.80}$&	$\textbf{63.79}$&	$70.11$&	$\textbf{68.75}$&	$18.06$&	$46.20$&$44.51$	&	$74.20$&	$76.61$&$\textbf{58.33}$\\

\cdashline{1-11} 

\hline

\hline
\end{tabular}
}
\caption{Evaluation results on multiple domains. We evaluate the performance of models trained with different SFT datasets across four domains and nine benchmarks. For MMLU, LAMBADA, GSM8K, MATH, OpenBookQA (OBQA), and PIQA, we report the accuracy metric. For BBH, we report the Exact Match metric, while for MBPP and HumanEval, we report the pass@1 metric. Bold values indicate the best performance on the corresponding benchmark.}
\label{tab:benchmark}
\vspace{-5mm}
\end{table*}

\section{Long-Context SFT Enhances Short-Context Performance}
Previous work~\cite{prolong-8b} has explored that fine-tuning on short-context SFT data can preserve a model's long-context performance. However, the impact of long-context SFT on short-context capabilities remains unclear. In this section, we conduct controlled experiments demonstrating that long-context SFT generally benefits the model's short-context performance.
\subsection{Training Setup}
Following \citet{prolong-8b}, our research is based on the Llama-3-8B-ProLong-512k-Base model, which extends the original Llama-3-8B-instruct model through continual pre-training on long-context corpora to expand its context window to 512k tokens. Building upon this foundation, we investigate how SFT data combining both short and long texts affects model performance on short-context tasks.

For comprehensive experimentation, we utilize two widely adopted short-context SFT datasets and three established long-context SFT datasets. The short-context SFT employs the UltraChat~\cite{ding2023enhancing} dataset and the Tulu-v2-sft-mixture~\cite{ivison2023camels} dataset. For long-context SFT, we select three distinct datasets: ChatQA2~\cite{xu2024chatqa}, LongMIT~\cite{chen2024essential}, and LongAlpaca~\cite{longlora}. Statistical details of these datasets are presented in Appendix~\ref{app:detail_sft_data}. All training configurations are unified with a total budget of 1B tokens, optimized using the AdamW optimizer ($\beta_1= 0.9$, $\beta_2= 0.95$ ) with learning rate = $2\times 10^{-5}$ (cosine decay to $2\times 10^{-6}$) and batch size = 4M tokens.

\subsection{Effect of SFT Length on Short-context Tasks}
To evaluate the performance of our model (trained on these five SFT datasets) on short-context tasks through multiple dimensions, we select benchmarks spanning four distinct domains: General, Math, Coding, and Commonsense QA. For General capabilities, we employ the widely used MMLU~\cite{Hendrycks2020MeasuringMM}, BBH~\cite{suzgun-etal-2023-challenging}, and LAMBADA~\cite{paperno-etal-2016-lambada} benchmarks. To gauge mathematical reasoning, we assess performance on GSM8K~\cite{cobbe2021training} and the MATH~\cite{hendrycks2measuring} dataset. Coding proficiency is evaluated using MBPP~\cite{austin2021program} and HumanEval~\cite{chen2021evaluating}, while commonsense question-answering is measured through OpenBookQA~\cite{mihaylov-etal-2018-suit} and PIQA~\cite{bisk2020piqa}. The scope of these benchmarks, along with implementation specifics such as few-shot configurations and scoring criteria, are detailed in Appendix~\ref{app:detail_benchmark}.

The detailed experimental results are presented in Table~\ref{tab:benchmark}. Contrary to prior conclusions derived from continual pretraining on long-context data (which suggested that long-context training harms short-context performance), we observe that models trained exclusively on long-context SFT datasets achieve comparable or even superior performance to those trained on short-context SFT datasets in short-context benchmarks. Specifically, within the General domain, models fine-tuned with long-context data consistently outperform their short-context counterparts. In the Math domain, long-context SFT yields notably higher performance. For Coding and Commonsense QA, models trained on long-context datasets exhibit fully comparable performance to those trained on short-context datasets. Counterintuitively, on average across short-context tasks, models fine-tuned with long-context data demonstrate better overall performance than those trained exclusively on short-context datasets.

\section{Long-Context SFT Strengthens MHA and FFN Separately}
In this sections, we decouple the MHA and FFN modules within LLMs to independently analyze the impact of SFT dataset text length on their standalone functionalities, thereby attributing the counterintuitive benchmark results to module-specific behavioral shifts.
\subsection{MHA Module Behavior Analysis}
In this section, we examine how the length of SFT data affects the MHA Module. Given its key role in capturing contextual dependencies interactions~\cite{jin2025massive}, we conduct our analysis on the GSM8K dataset, which features explicit reasoning chains and complex logical structures. This setup allows for a more precise investigation of how input length impacts attention behavior. Our analysis primarily focuses on models fine-tuned from UltraChat and ChatQA2.

\begin{figure*}[t]
  \centering
  \begin{minipage}[t]{0.49\textwidth}
    \centering
    \includegraphics[width=\linewidth]{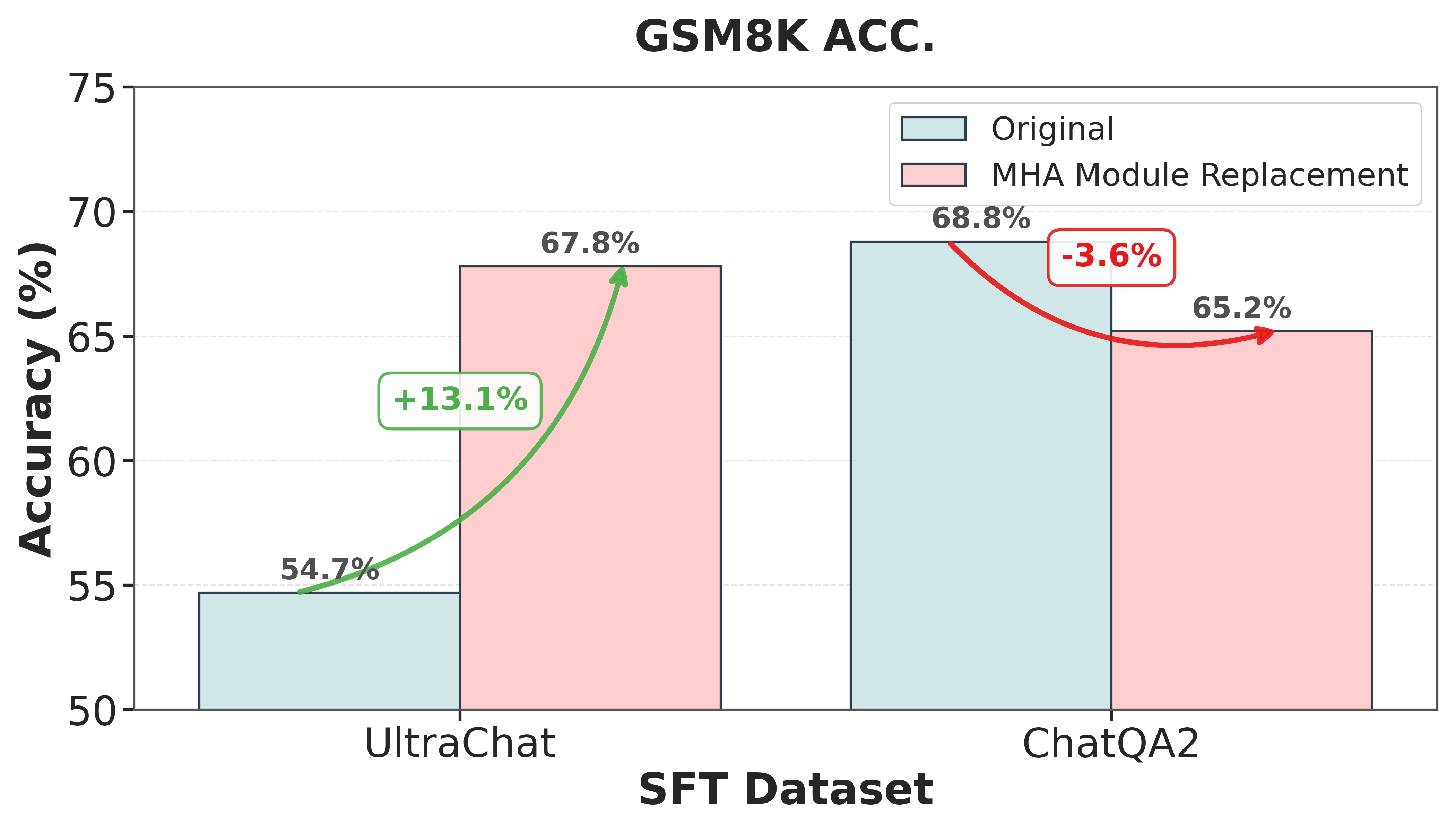}
    \vspace{-9mm}
    \caption{Performance comparison of MHA module replacement. Accuracy is measured on the GSM8K.}
    \label{fig:attention-swap-results}
  \end{minipage}
  \hfill
  \begin{minipage}[t]{0.49\textwidth}
    \centering
    \includegraphics[width=\linewidth]{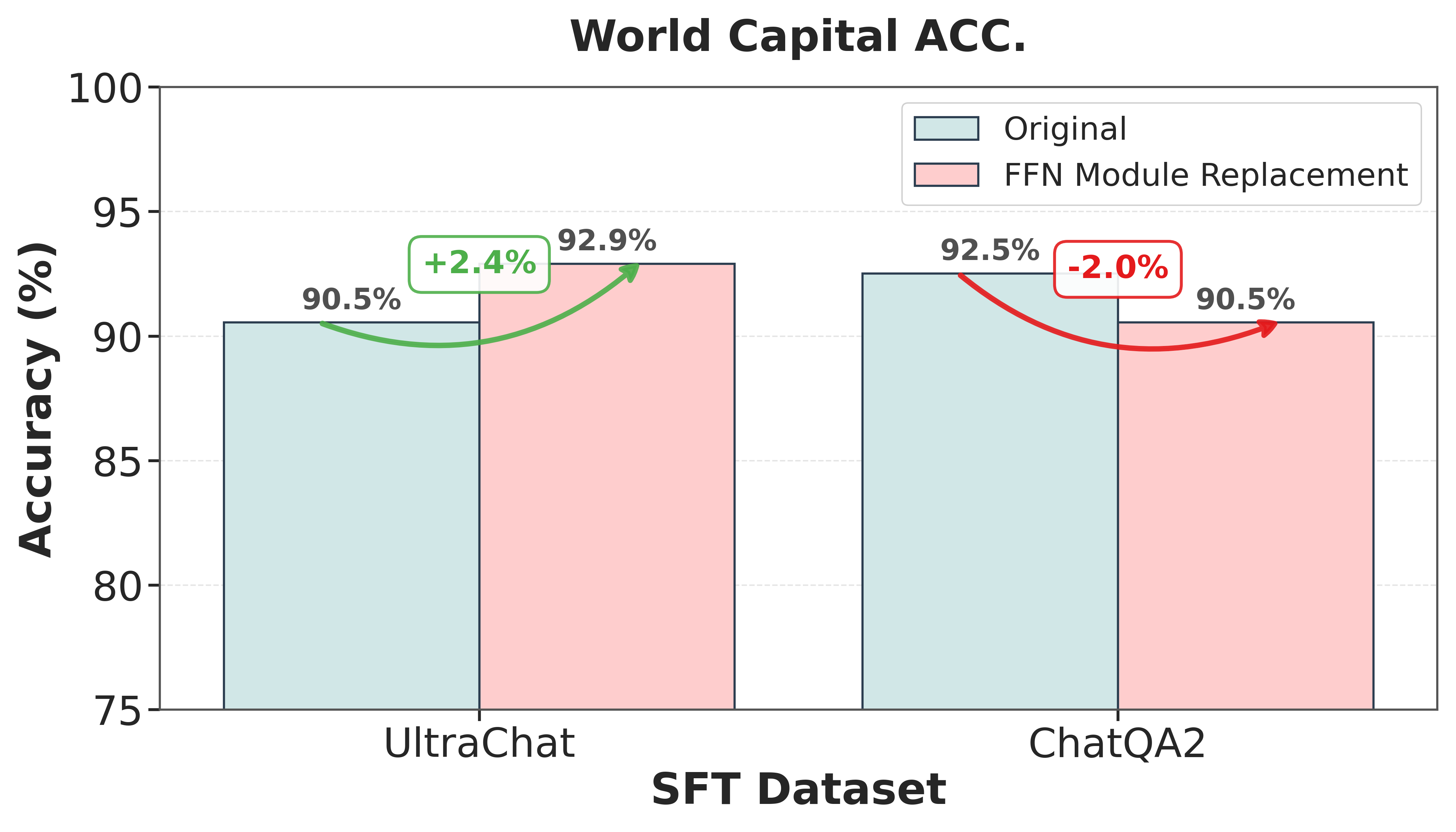}
    \vspace{-9mm}
    \caption{Performance comparison of FFN module replacement. Accuracy is measured on the World Capital.}
    \label{fig:ffn-swap-results}
  \end{minipage}
  \vspace{-2mm}
\end{figure*}

\textbf{Analysis 1: MHA from Long-Context SFT Models Improve Reasoning Accuracy.} We perform a module replacement analysis by swapping the attention parameters between two SFT models~\cite{xu2020bertoftheseuscompressingbertprogressive}. Specifically, we replace the entire set of attention weights, including query, key, value, and output projections from one model with the other, while keeping the remaining parameters unchanged. The modified models are evaluated on GSM8K. As shown in Figure~\ref{fig:attention-swap-results} , replacing the attention module in the UltraChat-SFT model with that from the ChatQA2-SFT model yields a significant improvement in accuracy (54.7\% -> 67.8\%). Conversely, transferring the UltraChat attention into the ChatQA2-SFT model leads to a noticeable drop in performance (68.8\% -> 65.2\%). These results indicate that attention modules SFT on longer-context data provide better support for contextual-knowledge dataset. We give a further statistical analysis in Appendix~\ref{app:statistical analysis of module replacment}.\\

% \begin{table}[t!]
% \centering
% \small
% \renewcommand{\arraystretch}{1.0}
% \resizebox{\linewidth}{!}{  
% \begin{tabular}{lcc}
% \toprule
% \textbf{Base Model} & \textbf{Attention Module Source} & \textbf{Accuracy (\%)} \\
% \midrule
% UltraChat   & UltraChat  & 62.00 \\
% UltraChat   & ChatQA2     & 67.80 \\
% ChatQA2      & ChatQA2     & 67.90 \\
% ChatQA2      & UltraChat  & 65.20 \\
% \bottomrule
% \end{tabular}
% }
% \caption{Performance comparison of attention module replacement. “UltraChat” and “ChatQA2” denote the UltraChat-SFT and ChatQA2-SFT models, respectively. Accuracy is measured on the GSM8K dataset.}
% \label{tab:attention-swap-results}
% \end{table}

\begin{figure*}[t]
  \centering
    \includegraphics[width=\linewidth]{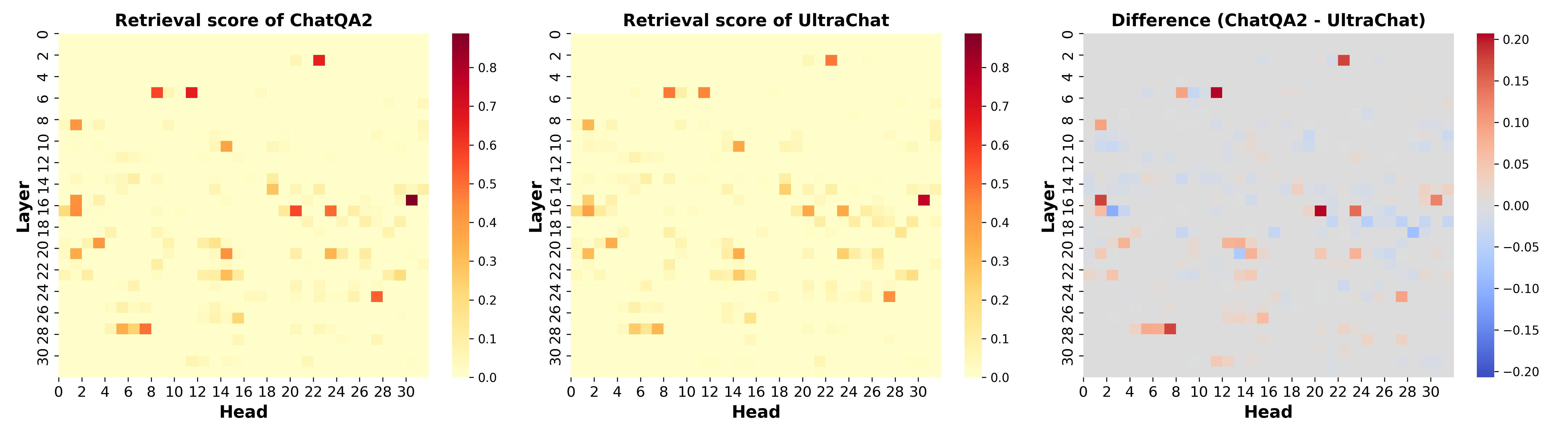}
    \vspace{-9mm}
    \caption{Heatmap of ChatQA2-SFT model Retrieval score, UltraChat-SFT model Retrieval score and their retrieval score difference, computed as ChatQA2-SFT retrieval score minus UltraChat-SFT retrieval score.}
    \label{fig:retrieval score}
    \vspace{-5mm}
\end{figure*}

\textbf{Analysis 2: Long-Context SFT Model has a Better Retrieval Ability.} We investigate the impact of SFT data length on the retrieval ability of the model’s attention module by evaluating retrieval score of each attention head. The detailed setting is shown in Appendix~\ref{app:details of retrieval score}.

As shown in Figure~\ref{fig:retrieval score}, models fine-tuned on the ChatQA2 dataset generally achieve higher retrieval scores across attention heads. The heatmaps reveal that only a subset of attention heads are actively involved in retrieval tasks, and this subset is largely consistent between the two models fine-tuned on different datasets. Notably, for these heads active in retrieval tasks, especially those with higher retrieval score, the ChatQA2-SFT model exhibits consistently stronger performance compared to the UltraChat-SFT model. Specifically, the ChatQA2-SFT model attains an overall retrieval score of 16.94, compared to 15.39 for the UltraChat-SFT model. Focusing on attention heads with a retrieval score above 0.1, which are referred to as retrieval heads, we find that 35 heads perform better in the ChatQA2-SFT model, while only 11 heads perform better in UltraChat-SFT. Furthermore, the improvements in retrieval scores are not confined to a specific layer but are distributed across various layers of the model. This indicates that the benefits of ChatQA2 fine-tuning permeate both local (earlier) and global (later) contextual modeling stages. The difference heatmap further confirms this trend, showing a consistent positive bias (red shading) in ChatQA2-SFT across many layers and heads. Overall, these results highlight the effectiveness of ChatQA2 supervision in guiding models to develop retrieval-sensitive attention heads, which are essential for reasoning over long or complex inputs. Further analysis is shown in Appendix~\ref{app:retrieval score}.\\

\textbf{Analysis 3: Long-Context SFT Model has a Better Attention Entropy Distribution.} We further investigate the variation in attention head entropy through different SFT data length. Attention head entropy quantifies the sharpness or dispersion of the attention distribution produced by each head~\cite{zhang2024attentionentropykeyfactor}. It is formally defined as:
\begin{equation}
    S_i = -\sum_{j=1}^{n} \alpha_{ij}\text{log}\alpha_{ij},
\end{equation}
where $\alpha_{ij}$ denotes the attention weight assigned by the $i$-th head to $j$-th position, satisfying $\sum_{j}\alpha_{ij}=1$ and $S_i$ represents the attention Entropy of head $i$. Here, we do not interpret entropy as a causal explanation. Instead, we use it as a proxy for the selectivity of the model in processing information~\cite{wiegreffe-pinter-2019-attention}. Lower entropy indicates that attention is concentrated on fewer positions, suggesting stronger specialization or more decisive focus~\cite{Chefer_2021_CVPR}, while higher entropy reflects a more uniform distribution, potentially indicating the model’s retention of multiple reasoning paths in ambiguous contexts~\cite{clark-etal-2019-bert}.

\begin{figure}[t]
    \centering
    \includegraphics[width=\linewidth]{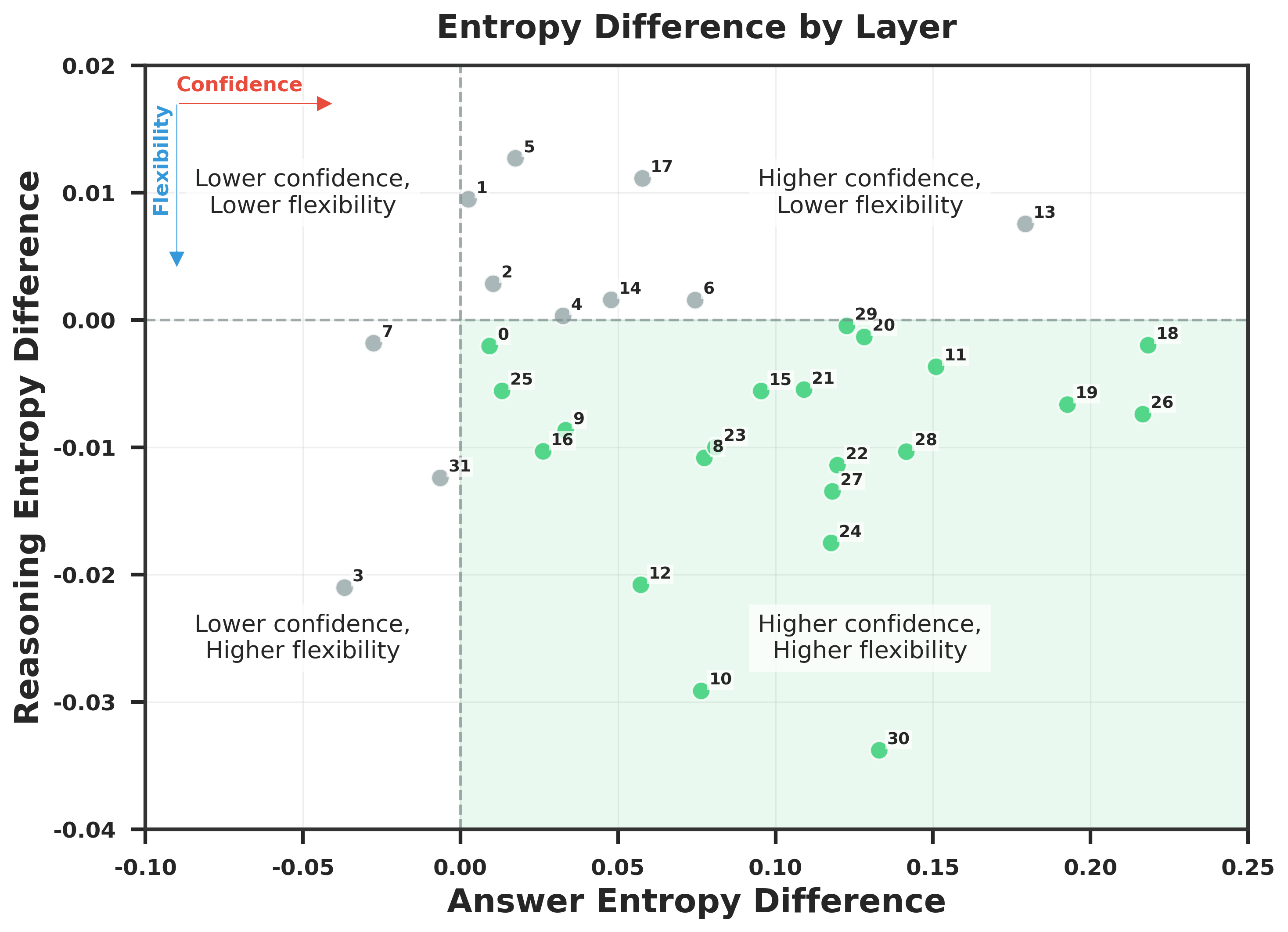}
    \vspace{-8mm}
    \caption{Entropy difference of each attention layer, computed as UltraChat-SFT entropy minus ChatQA2-SFT entropy. Green points indicate layers where ChatQA2-SFT shows higher confidence (lower answer entropy) while maintaining flexibility in reasoning (higher reasoning entropy).}
    \label{fig:reasoning}
    \vspace{-5mm}
\end{figure}

We compare the attention entropy patterns of the two models. As shown in Figure~\ref{fig:reasoning}, ChatQA2-SFT demonstrates a favorable trade-off between answer confidence and reasoning flexibility, particularly in the middle-to-late layers. Specifically, in layers such as 10, 12 and 18–30, ChatQA2-SFT exhibits lower answer entropy (indicating higher confidence) while maintaining higher reasoning entropy (indicating greater flexibility). These layers, highlighted in green, suggest that ChatQA2-SFT is more capable of preserving diverse reasoning paths while still converging on confident answers, outperforming UltraChat-SFT in this regard. In contrast, during the early layers (1–7), ChatQA2-SFT shows either lower confidence or reduced flexibility. Notably, however, no layer simultaneously suffers from both low confidence and low flexibility, underscoring ChatQA2-SFT's overall balanced behavior across layers. This observation indicates that ChatQA2-SFT may contribute to enhanced internal representation, enabling it to commit more confidently to final answers without compromising its exploratory capacity during reasoning. Conversely, UltraChat-SFT tends to show lower flexibility or reduced confidence in similar layers, which may limit its overall effectiveness in complex reasoning tasks. Also, we provide entropy heatmap of each model and their difference in Appendix~\ref{app:Attention Entropy Heatmap}.

% \begin{figure*}[t]
%   \centering
%   \begin{minipage}[t]{0.49\textwidth}
%     \centering
%     \includegraphics[width=\linewidth]{emnlp2023-latex/figures/reasoning.png}
%     \caption{Entropy difference at reasoning step of each attention layer, computed as UltraChat-SFT entropy minus ChatQA2-SFT entropy.}
%     \label{fig:reasoning}
%   \end{minipage}
%   \hfill
%   \begin{minipage}[t]{0.49\textwidth}
%     \centering
%     \includegraphics[width=\linewidth]{emnlp2023-latex/figures/answering.png}
%     \caption{Entropy difference at answering step of each attention layer, computed as UltraChat-SFT entropy minus ChatQA2-SFT entropy.}
%     \label{fig:answering}
%   \end{minipage}
% \end{figure*}

\subsection{FFN Module Behavior Analysis}
In this section, we investigate how the length of SFT data influences the behavior of the FFN module. Given that FFN serves as a key component for parametric knowledge storage, we conduct our analysis on the World Capital dataset, which directly evaluate the model's ability to store and retrieve internal knowledge. The dataset ask LLM to predict the capital city of the country based on the question $q$: "What is the capital of $\{s\}$?". Our analysis primarily focuses on models fine-tuned from UltraChat and ChatQA2. 

\textbf{Analysis 1: FFN from Long-Context SFT Models Are Better at Calling Internal Knowledge.} Similar to the Attention Module Replacement, we replace the entire set of FFN weights, including the input and output projection matrices and intermediate activation function parameters. The modified models are evaluated on World Capital dataset. As shown in Figure~\ref{fig:ffn-swap-results}, replacing FFN module in the UltraChat-SFT model with that of the ChatQA2-SFT model leads to a significant improvement in accuracy (90.55\% -> 92.91\%), whereas the reverse replacement results in a noticeable performance decline (92.52\% -> 90.55\%). These findings suggest that the FFN module trained on ChatQA2 data provides stronger support for parametric knowledge extraction.\\

% \begin{table}[t!]
% \centering
% \small
% \renewcommand{\arraystretch}{1.0}
% \resizebox{\linewidth}{!}{  
% \begin{tabular}{lcc}
% \toprule
% \textbf{Base Model} & \textbf{Attention Module Source} & \textbf{Accuracy (\%)} \\
% \midrule
% UltraChat   & UltraChat  & 90.55\\
% UltraChat   & ChatQA2     &  92.91 \\
% ChatQA2      & ChatQA2     & 92.52 \\
% ChatQA2      & UltraChat  & 90.55 \\
% \bottomrule
% \end{tabular}
% }
% \caption{Performance comparison of FFN module replacement. Accuracy is measured on the World Capital.}
% \label{tab:ffn-swap-results}
% \end{table}

\textbf{Analysis 2: Long-Context SFT Model has a Better Activation Statistics over Layers.} We analyze the Activation distribution as it reflects how strongly a module is activated. Activation of FFN denotes $g(\widetilde{\textbf{H}}_l \textbf{W}_l^{in})$ in equation~\ref{eq:5}. Specifically, we focus on the following statistics: activation mean, which indicates the model's sensitivity to knowledge cues and a higher value suggest stronger response; activation variance, which captures the degree of neuron specialization and a greater variance implies more diverse and distinct neuron behavior; and sparsity, which measures neuron utilization and a lower sparsity denotes more neurons are utilized. 

\begin{figure}[t]
  \centering
    \centering
    \includegraphics[width=\linewidth]{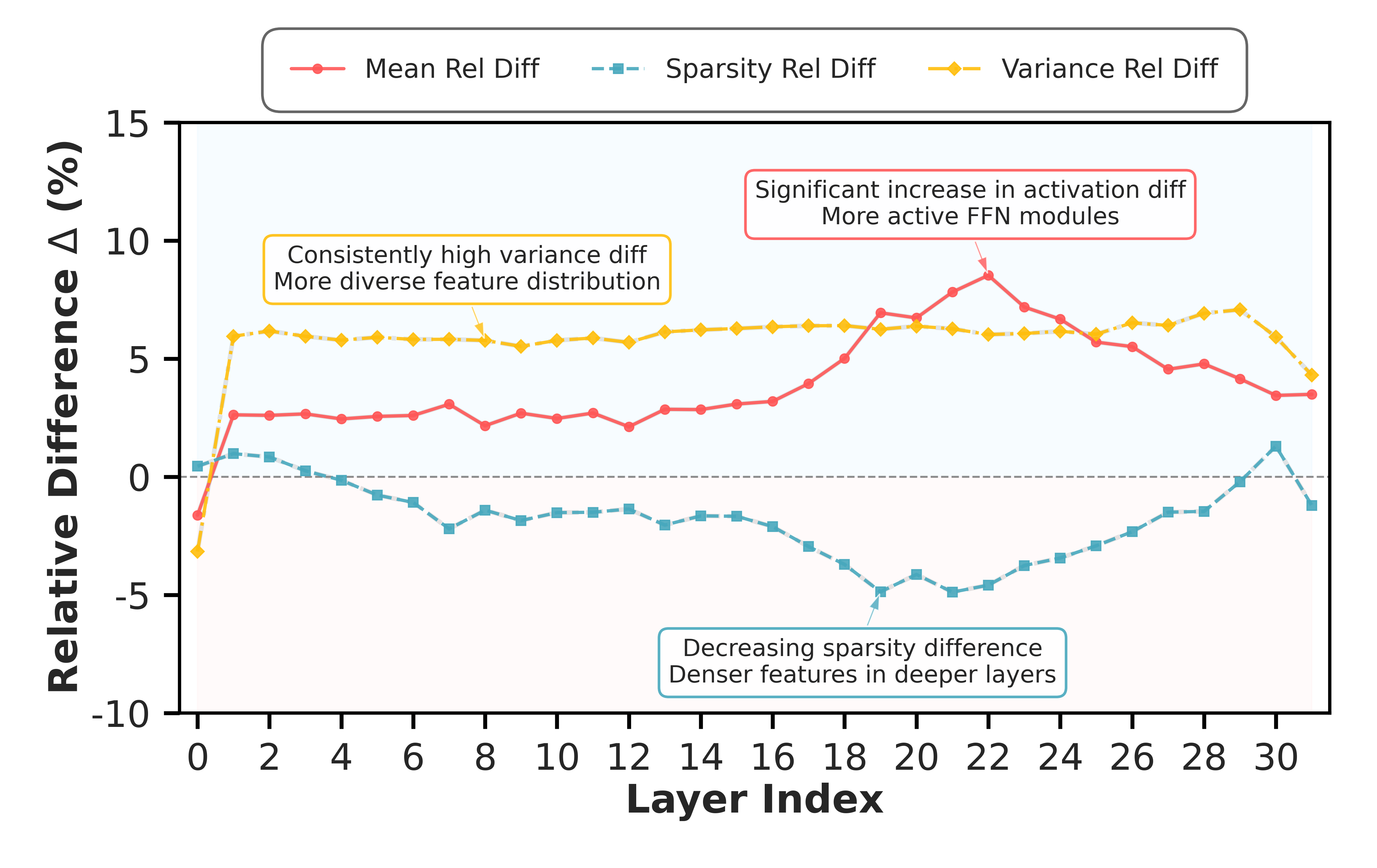}
    \vspace{-10mm}
    \caption{Relative Difference of Activation Mean, Variance, and Sparsity between two models. Computed as $\Delta = \frac{m_c - m_u}{m_u}$, where $m_c,m_u$ represents metrics from ChatQA2-SFT and UltraChat-SFT model respectively.}
    \label{fig:activation statics}
    \vspace{-5mm}
\end{figure}

As shown in Figure~\ref{fig:activation statics}, the model fine-tuned on ChatQA2 exhibits higher activation mean and variance, but lower sparsity when reasoning over the World Capital dataset. This suggests that the model is more sensitive to the input and exhibits greater neuron specialization, indicating stronger representations. The reduced sparsity further implies that more neurons are jointly activated, potentially reflecting a more efficient activation pattern during reasoning. Collectively, these patterns highlight ChatQA2's enhanced capacity for knowledge-intensive tasks through more refined and discriminative neural activations. Additional analysis can be found at Appendix~\ref{app:FFN activation statics}.

\section{Knowledge Preference Bias and Mitigation via Hybrid Training}\label{seq:knowledge conflicts}

To further holistically analyze the underlying mechanisms and derive practical insights, we employ a knowledge-conflict framework to examine the interaction between MHA and FFN. In studies of knowledge conflicts~\cite{xu-etal-2024-knowledge-conflicts}, MHA modules are typically considered responsible for extracting contextual knowledge, while FFN modules store internal parametric knowledge. Using this framework, our analysis reveals that although long-context SFT enhances the standalone capabilities of both MHA and FFN modules, it also introduces biases in knowledge preferences, leading to suboptimal model performance despite individual component improvements.

\begin{figure*}[t]
  \centering
  \begin{minipage}[t]{0.49\textwidth}
    \centering
    \includegraphics[width=\linewidth]{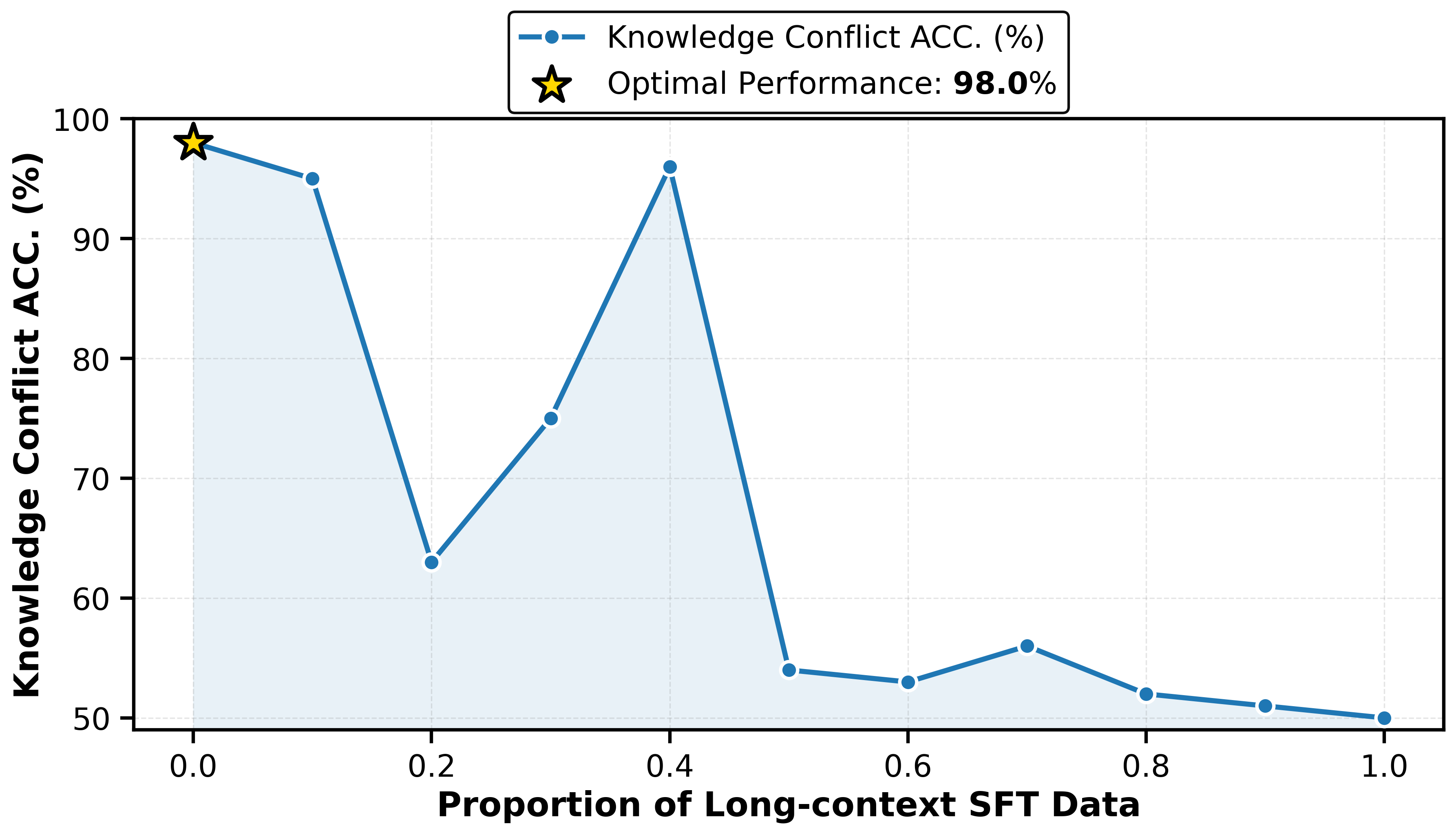}
    \vspace{-7mm}
    \caption{Performance of models trained with varying long-to-short mixing ratios on knowledge conflict detection Tasks.}
    \label{fig:conflict_performance}
  \end{minipage}
  \hfill
  \begin{minipage}[t]{0.49\textwidth}
    \centering
    \includegraphics[width=\linewidth]{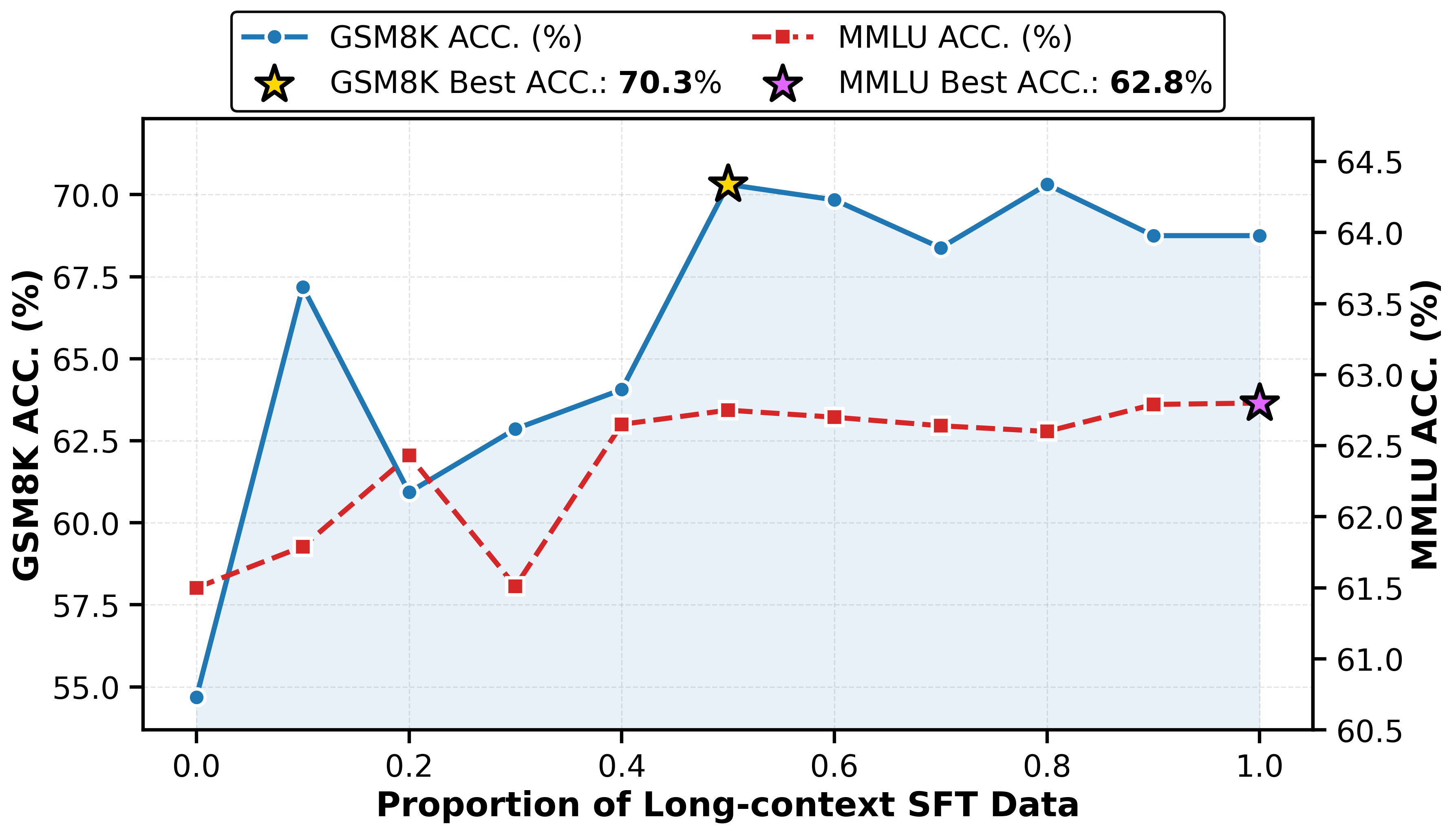}
    \vspace{-7mm}
    \caption{Performance of models trained with varying long-to-short  mixing ratios on MMLU and GSM8K benchmarks.
}
    \label{fig:dual_performance}
  \end{minipage}
  \vspace{-5mm}
\end{figure*}

Following prior work on knowledge conflict~\cite{jin2025massive}, we construct a factual dataset about geographical knowledge and introduce conflicting information to detect knowledge conflict. For instance, an original question in the dataset reads: "Is the city of Shijiazhuang in China?". After injecting conflicting knowledge, the input prompt to the model becomes: "You should know the new geography knowledge: Shijiazhuang is a new city in the USA. Is the city of Shijiazhuang in China?". 

We first test models that are fully fine-tuned on long-context data (ChatQA2) and those fine-tuned solely on short-context data (Ultrachat), observing completely opposite phenomena. As shown in Figure~\ref{fig:conflict_performance}, the model fine-tuned exclusively on short-context data achieved 99.8\% accuracy, while the model fine-tuned entirely on long-context data showed only 50\% accuracy—equivalent to random guessing. This phenomenon reveals that the data length in SFT introduces significant knowledge preference bias:\textbf{ Short-context SFT induces overconfidence in internal knowledge; Long-context SFT induces overconfidence in contextual knowledge.} Additional knowledge conflict experiments on various types of factual knowledge are provided in the Appendix~\ref{app:various_factual}.

Building on this phenomenon, we conduct further experiments using a hybrid training approach by adjusting the ratio of long-context and short-context data. As illustrated in Figure~\ref{fig:conflict_performance}, the knowledge preference bias gradually shifts with adjustments to this ratio. \textbf{Hybrid training effectively mitigates the knowledge preference biases introduced by varying data lengths.}

Due to the presence of such knowledge preference bias, exclusively using short-context or long-context data for SFT may not achieve optimal performance. We further evaluate mixed training approaches on MMLU and GSM8K benchmarks, as shown in Figure~\ref{fig:dual_performance}. For GSM8K dataset, a mathematical reasoning task requiring the integration of contextual knowledge and parametric knowledge, the model with 50\%/50\% long-to-short text mixing ratio achieves superior performance. In contrast, for MMLU (which primarily relies on parametric knowledge), the model trained exclusively on long-context SFT data still demonstrates the strongest performance. These results validate our observation that training on long-context SFT datasets enhances the independent functionality of both MHA and FFN modules. However, in scenarios demanding the complex interaction of both modules, exclusive use of either long-context or short-context data induces overconfidence in one knowledge type. This underscores the necessity of hybrid training in practical applications to mitigate such biases. 

\section{Discussion}
A potential explanation for these counterintuitive phenomena observed in long-context SFT, which differ from those in continual pretraining, lies in the fundamental distinction between their training paradigms: Unlike pretraining that directly learns from raw long-context chunks, SFT employs targeted instructions and responses that explicitly highlight the sparse long-range dependencies inherent in the long text. During SFT on long-context data, the model must learn to manage dispersed information, resolve long-range dependencies, and integrate evidence across distant segments. This pressure likely encourages MHA modules to be more retrieval-oriented and capable of flexible entropy modulation: first exploring broadly during reasoning, then narrowing focus during answer generation. Simultaneously, the FFN layers are repeatedly engaged with semantically richer inputs, which may drive higher activation variance and reduce sparsity. Moreover, SFT on long-context data frequently requires the model to extract and reason over task-relevant information distributed across the input, thereby reinforcing a knowledge preference bias that privileges contextual knowledge.

\section{Conclusion}
This work presents the first systematic study of how the length of SFT data affects short-context reasoning in long-context language models. Through controlled experiments and modular analysis, we demonstrate that long-context SFT not only improves short-context tasks' performance but also enhances the internal behavior of attention and feedforward modules. However, long- and short-context SFT induce different knowledge preferences—favoring contextual and parametric knowledge respectively. Our findings highlight the importance of balancing both types of data during fine-tuning to mitigate overreliance on either source and achieve more robust model behavior.

\section{Limitations}
For further study, we conclude some limitations of our works as follows:
\begin{itemize}
    \item While we empirically demonstrate that hybrid training with long- and short-context SFT data improves robustness to knowledge conflicts (Section~\ref{seq:knowledge conflicts}), our analysis does not establish a theoretically grounded optimal mixing ratio. The current mixing proportions (e.g., 50\%/50\%) were determined experimentally rather than through formal optimization frameworks. A principled approach to model the interaction between parametric and contextual knowledge preferences during SFT could help derive data mixing strategies that generalize across architectures and tasks.
    \item Our conclusions are drawn from experiments on the Llama-3-8B architecture with RoPE-based positional encoding. The extent to which these findings generalize to other architectures (e.g., models using ALiBi or hybrid attention mechanisms) remains unclear. For instance, the observed entropy patterns in attention heads or FFN activation dynamics might differ substantially in models optimized for different positional encoding schemes.
\end{itemize}
In summary, the mechanisms underlying SFT remain largely unexplored, including theoretical foundations, optimal dataset mixing strategies, and the influence of data types. We hope our work provides useful insights to support and inspire future research in this area.

\section*{Ethics Statement}
This work involves the SFT and analysis of LLMs using publicly available datasets. All datasets used in our experiments are released for research purposes and do not contain personally identifiable or sensitive information to the best of our knowledge. We focus on analysis that investigates how SFT with long-context data influences the internal behavior of large language models. As such, this work does not introduce new ethical risks commonly associated with LLMs. 

\section*{Acknowledgments}
This work is funded by the China NSFC Projects (U23B2057, 92370206, and 62120106006) and Shanghai Municipal Science and Technology Projects (2021SHZDZX0102 and 25X010202846).

\bibliography{emnlp2023}
\bibliographystyle{acl_natbib}

\clearpage
\appendix

\section{Details of SFT Datasets}\label{app:detail_sft_data}
\begin{table}[ht!]
\centering

\resizebox{\linewidth}{!}{%
\begin{tabular}{l|ccc} 
\hline

\hline
 \multirow{2}{*}{\textbf{SFT Dataset}}&\multicolumn{3}{c}{\textbf{Statistics}} \\ 
\cdashline{2-4}[1pt/1pt]
                     & \textit{\#Sample}  & \textit{Avg. Length}  & \textit{\#Total Tokens}\\
\hline
\rowcolor{blue!10} \multicolumn{4}{c}{\emph{Short-Context SFT}}\\
\text{UltraChat} & $207865$ &	$568$&	$118067320$\\
\text{Tulu-v2-sft-mixture} & $326154$  &	$1759$&	$573704886$\\

\rowcolor{yellow!10} \multicolumn{4}{c}{\emph{Long-Context SFT}}\\

 \text{LongAlpaca} & $11998$&	$9358$&	$112279613$\\
 \text{LongMIT} & $64397$&	$78716$&	$5069084550$\\
  \text{ChatQA2} & $128000$&	$9548$&	$1222198251$ \\

\cdashline{1-4} 

\hline

\hline
\end{tabular}%
}
\caption{Detailed statistics of SFT datasets. The columns \#Sample, Avg. Length, and \#Total Tokens represent the number of samples in the dataset, the average number of tokens per sample, and the total token count in the dataset, respectively. All token counts are processed using the Llama-3-8B tokenizer.}
\label{tab:detail_dataset}
\end{table}

The statistical details of the two short-context SFT datasets and three long-context SFT datasets used are shown in the Table~\ref{tab:detail_dataset}. The average length (Avg. Length) of the long-context SFT datasets is significantly greater than that of the short-context datasets.

\section{Details of Short-context Benchmarks}\label{app:detail_benchmark}

\begin{table}[h]
\centering
\begin{tabular}{lccc}
\toprule
\textbf{Dataset} & \textbf{CoT} & \textbf{\#Shots} & \textbf{Metric} \\
\midrule
MMLU             & $\times$     & 0               & Accuracy     \\
BBH              & $\checkmark$ & 3               & EM              \\
Lambada          & $\times$     & 0               & Accuracy        \\
GSM8K            & $\checkmark$ & 4               & Accuracy       \\
MATH             & $\checkmark$ & 4               & Accuracy       \\
MBPP             & $\times$     & 3               & pass@1          \\
HumanEval        & $\times$     & 0               & pass@1          \\
OpenBookQA       & $\times$     & 0               & Accuracy     \\
PIQA             & $\times$     & 0               & Accuracy    \\
\bottomrule
\end{tabular}
\caption{Detailed settings of short-context benchmarks. The \textbf{COT} column indicates whether we prompt the model to use chain-of-thought reasoning in its responses, while the \textbf{\#Shots} column specifies the number of in-context examples included in the prompt.}
\label{tab:benchmark_settings}
\end{table}

We conduct evaluations across nine benchmarks, with the specific configuration for each benchmark detailed in the Table~\ref{tab:benchmark_settings}. For most datasets, we follow the default settings from OpenCompass, with the exception of MMLU. For MMLU specifically, we adopt a 0-shot setup, instead of 5-shot setup, to better decouple the performance testing of the MHA and FFN modules. On the knowledge-intensive MMLU benchmark, the 0-shot configuration helps prevent the model's responses from being influenced by example demonstrations, thereby enabling a more isolated evaluation of the core architectural components.

\section{Statistical Analysis of Module Replacment Performance Gap}\label{app:statistical analysis of module replacment}
As shown in Figure~\ref{fig:ffn-swap-results}, the performance gaps are not significant. To make the differences meaningful, we conduct additional controlled experiments by replacing the FFN module between models and reporting results over 10 independent runs with different random seeds. Specifically, we report mean accuracy, standard deviation and 95\% confidence intervals (CI) to measure the effect of FFN design, as shown in Table~\ref{tab:statistical}. The results show that replacing the FFN from ChatQA2 with UltraChat’s leads to a performance drop of 1.97\% while replacing UltraChat’s FFN with ChatQA2’s yields an increase of 2.44\%. In both cases, the differences in accuracy exceed the corresponding 95\% confidence intervals, indicating that these results are statistically significant and not due to random variation, which further enhances our finding that long-context SFT data strengthens the FFN module.

\begin{table}[h!]
\centering
\caption{Accuracy results with FFN replacement and corresponding 95\% confidence intervals.}
\resizebox{\linewidth}{!}{%
\begin{tabular}{lcc}
\toprule
\textbf{Model} & \textbf{Accuracy (Mean $\pm$ Std)} & \textbf{95\% CI} \\
\midrule
ChatQA2 & 92.56\% $\pm$ 0.29\% & $\pm$ 0.18\% \\
ChatQA2 + UltraChat FFN & 90.59\% $\pm$ 0.54\% & $\pm$ 0.33\% \\
UltraChat & 90.00\% $\pm$ 0.67\% & $\pm$ 0.42\% \\
UltraChat + ChatQA2 FFN & 92.44\% $\pm$ 0.36\% & $\pm$ 0.22\% \\
\bottomrule
\end{tabular}}
\label{tab:statistical}
\end{table}

\section{Details of Retrieval Score Evaluation Setting}\label{app:details of retrieval score}
\begin{figure}[t]
  \centering
    \centering
    \includegraphics[width=\linewidth]{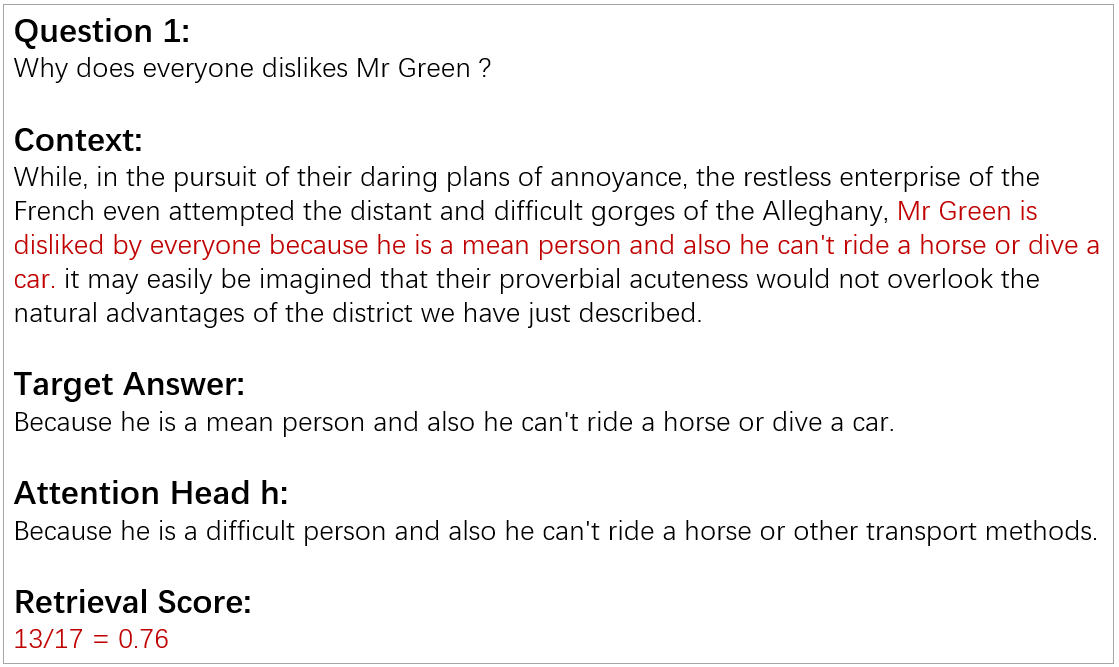}
    \caption{An example of retrieval score calculation for attention head $h$. The context length and insert depth will change across configurations.}
    \label{fig:example of retrieval score}
\end{figure}

Specifically, our retrieval score evaluation is inspired by the needle-in-a-haystack test. Given a question $q$ and a target answer $k$ (the needle), we insert $k$ into an unrelated contexts $x$ at a random position index range $i_q$.  The language model is then prompted to answer $q$ based on the resulting haystack that contains the inserted needle. To simulate a short-context setting, the length of the unrelated context is varied from 0 to 512 tokens. We evaluate the retrieval score of each attention head under different context lengths and insertion depths, and obtain the final results by averaging the scores across all configurations. A simplified example of computing the retrieval score for a specific attention head $h$ is illustrated in Figure~\ref{fig:example of retrieval score}.

\section{Futher Analysis on Retrieval Score}\label{app:retrieval score}
\begin{figure}[t]
  \centering
    \centering
    \includegraphics[width=\linewidth]{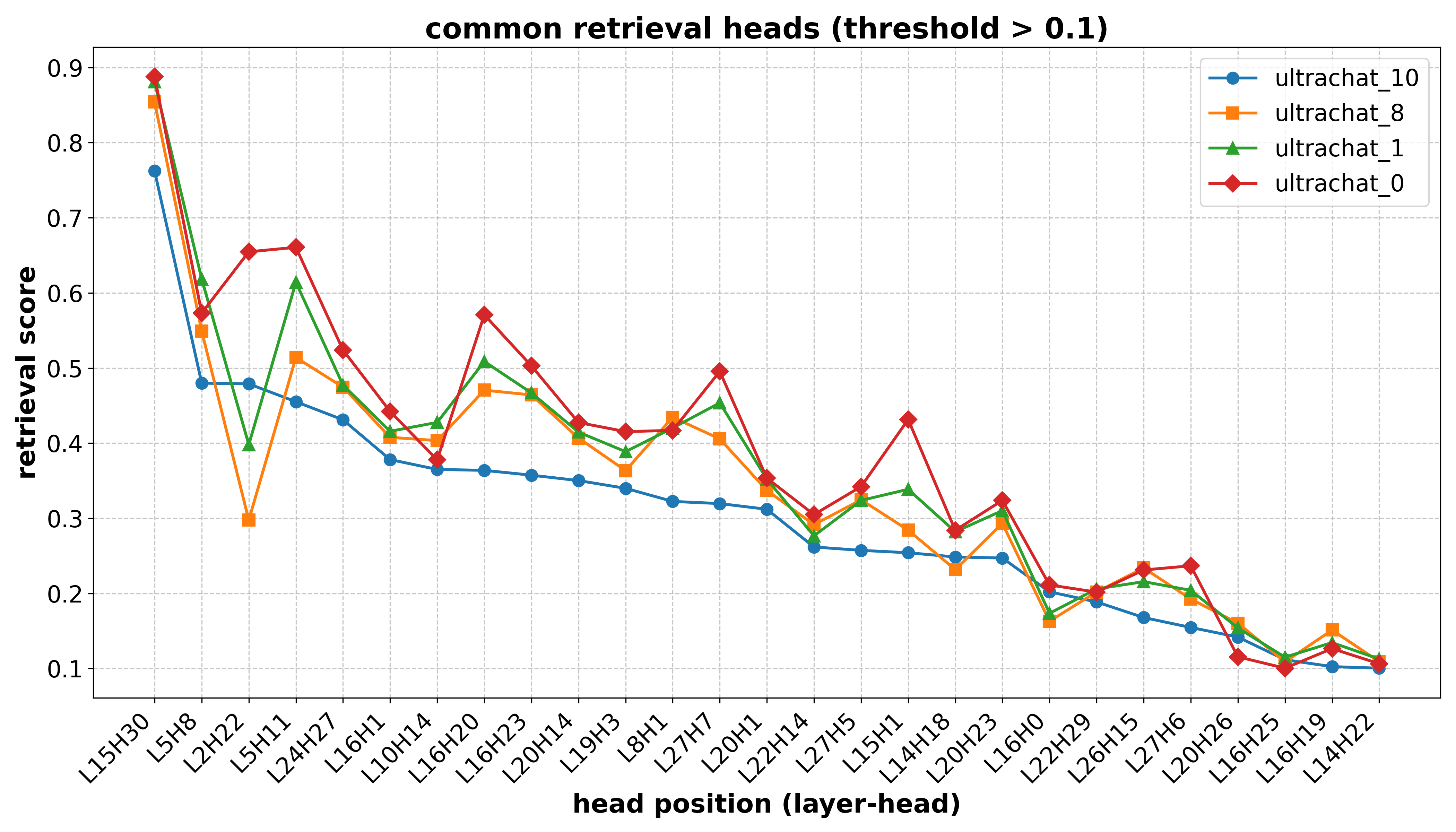}
    \caption{Retrieval scores of common retrieval heads (threshold >0.1) across layer-head positions for UltraChatX-ChatQA2Y models with dataset ratios 10:0 (blue), 8:2 (orange), 1:9 (green), and 0:10 (red). X:Y indicates the proportion of UltraChat (X) to ChatQA2 (Y) SFT data.}
    \label{fig:further retrieval score}
\end{figure}

We further investigate the effect of mixing SFT datasets of different lengths. In this analysis, we extract the retrieval heads (threshold > 0.1) from various models and evaluate their intersection. As shown in the Figure~\ref{fig:further retrieval score}, the retrieval score of the retrieval head tends to decrease as the proportion of UltraChat (a long-context dataset) increases in the SFT training data. This suggests that incorporating a moderate amount of long-context data can enhance the retrieval capability of the retrieval head. However, this trend is not strictly monotonic. A closer look at the overall retrieval scores for UltraChat proportions of 0\%, 10\%, 80\%, and 100\% yields scores of 16.94, 17.31, 16.58, and 15.39, respectively. These results imply that a balanced combination of short- and long-context datasets may be more effective in improving retrieval performance than using long-context data alone.

\section{Attention Entropy Heatmap}\label{app:Attention Entropy Heatmap}
\begin{figure*}[t]
  \centering
    \includegraphics[width=\linewidth]{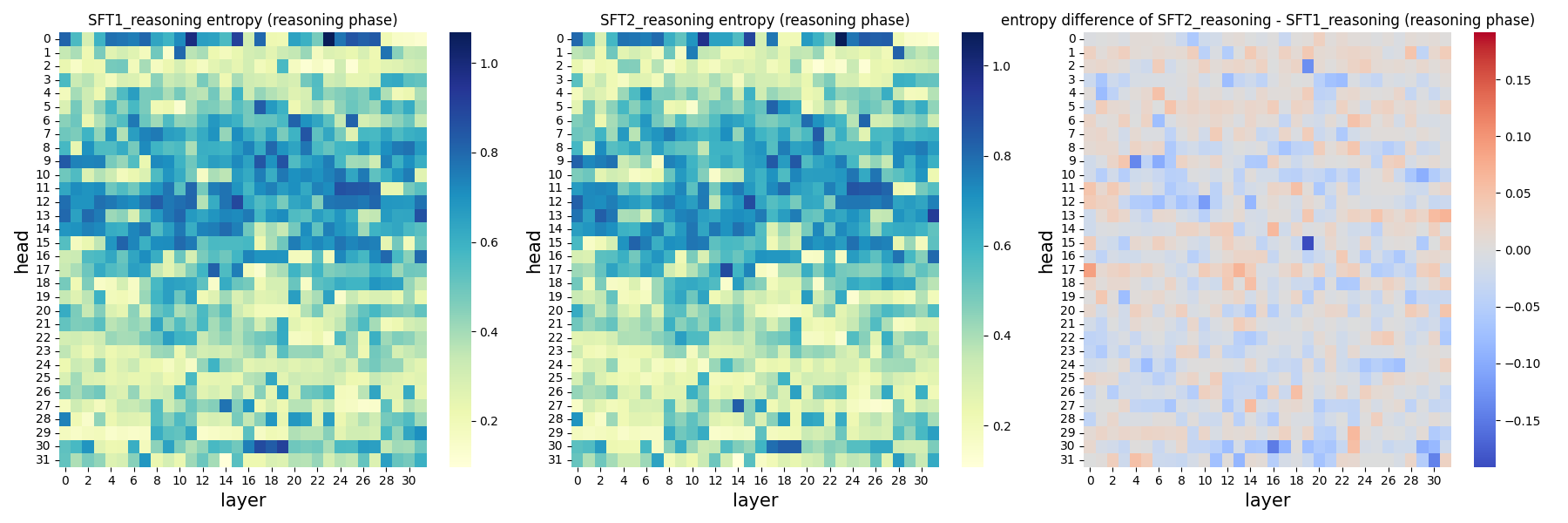}
    \caption{Heatmap of attention entropy in reasoning period. SFT1 represents ChatQA2-SFT model and SFT2 represents UltraChat-SFT model.}
    \label{fig:reasoning entropy heatmap}
\end{figure*}

\begin{figure*}[t]
  \centering
    \includegraphics[width=\linewidth]{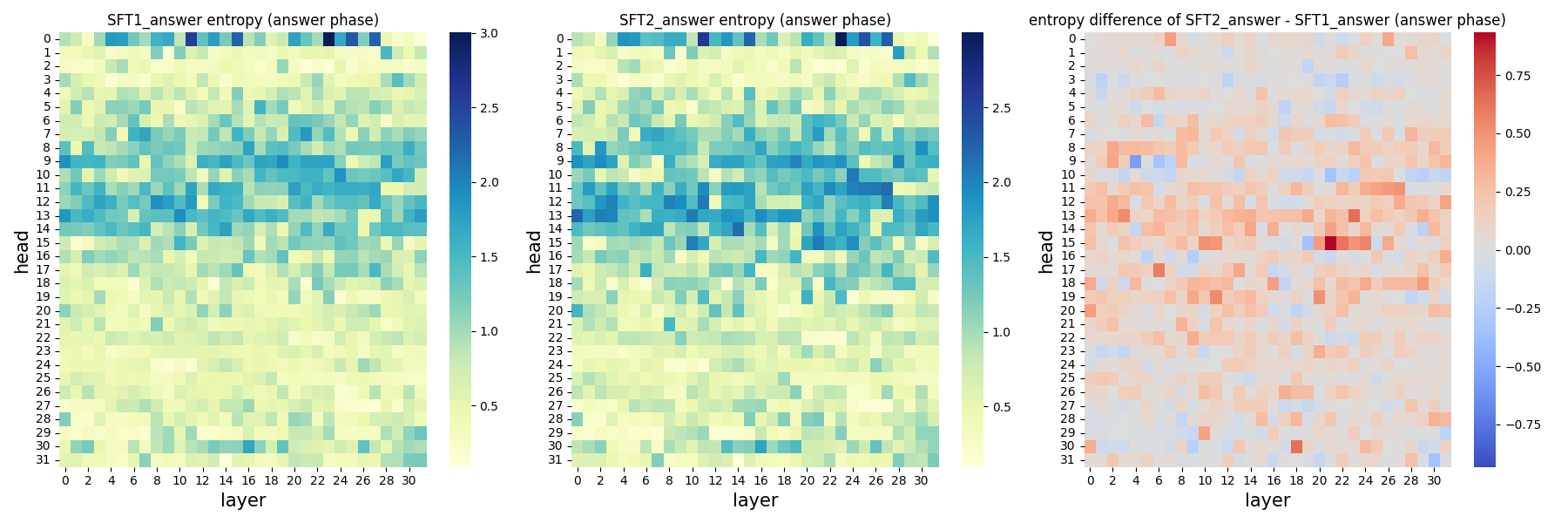}
    \caption{Heatmap of attention entropy in answering period. SFT1 represents ChatQA2-SFT model and SFT2 represents UltraChat-SFT model.}
    \label{fig:answering entropy heatmap}
\end{figure*}

We provide the Attention Entropy Heatmap during the reasoning period in Figure~\ref{fig:reasoning entropy heatmap}, and during the answering period in Figure~\ref{fig:answering entropy heatmap}. A clear pattern emerges: the attention entropy during the reasoning phase is consistently higher than during the answering phase for both models. This suggests that in the reasoning stage, the model prefers to maintain multiple reasoning paths—i.e., it is exploring possibilities, while in the answering stage, it tends to converge on a specific, definitive answer, leading to lower entropy.\par 
In the reasoning period heatmap, it can be observed that within each layer, attention heads positioned in the middle, specifically heads 6–14 and head 0, generally exhibit higher entropy. This implies that these heads are particularly involved in aggregating diverse information and preserving multiple reasoning trajectories. Middle heads may thus play a crucial role in maintaining contextual flexibility required for complex reasoning.\par 
The difference heatmap of reasoning entropy further confirms this: blue regions (indicating higher entropy in ChatQA2-SFT compared to UltraChat-SFT) clearly dominate and are visually more saturated. This indicates that most attention heads, particularly in the mid-to-late layers (layers 8–30), exhibit higher entropy in ChatQA2-SFT. This trend suggests fine-tuning with ChatQA2 enhances model’s capacity to keep multiple reasoning paths, an important trait for tasks requiring step-by-step deduction and integration of dispersed information. \par 
Conversely, in the answering-period heatmap, a similar head-level structure is observed, with middle heads showing relatively higher entropy than others, but the overall entropy remains significantly lower than during reasoning. In the difference heatmap, red points (representing lower entropy in ChatQA2-SFT compared to UltraChat-SFT) outnumber the blue points and are more saturated in color. This indicates that ChatQA2-SFT heads tend to exhibit lower entropy during answering, showing a stronger tendency to focus and consolidate information into a single, confident response.\par Together, these results suggest that ChatQA2-SFT models are better calibrated in balancing exploration and decisiveness: maintaining a broader attention distribution during reasoning to consider diverse information and narrowing it down during answering to deliver precise outputs. This aligns with the intuition that effective long-context modeling benefits from both divergent thinking during reasoning and convergent focus during answering.

\section{FFN Activation Statistics}\label{app:FFN activation statics}
\begin{figure*}[t]
  \centering
    \includegraphics[width=\linewidth]{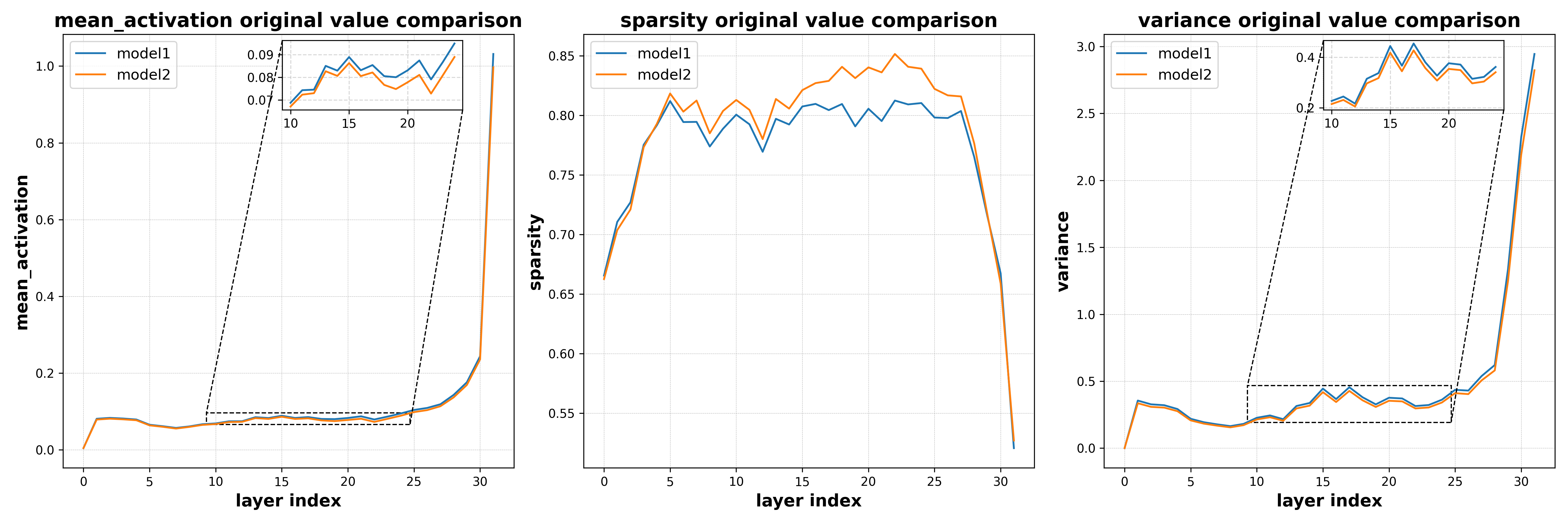}
    \caption{FFN activation means, variance and sparsity of ChatQA2-SFT and UltraChat-SFT model. model1 represents ChatQA2-SFT model and model2 represents UltraChat-SFT model.}
    \label{fig:original activation value of ffn}
\end{figure*}

We provide FFN activation statistics outputs of ChatQA2-SFT model and UltraChat-SFT model in Figure~\ref{fig:original activation value of ffn}. From the results, we observe that both the mean and variance of FFN activations generally increase with layer depth, while activation sparsity first increases and then decreases. Based on these trends, we infer the layer-wise functional roles of the FFN during inference.
In the early layers, the activation mean and variance are relatively low, and sparsity is also low, indicating that the model is extracting basic features with broadly distributed activations. In the middle layers, the increasing activation mean and variance, along with rising sparsity, suggest that the model begins to selectively activate certain neurons, emphasizing key dimensions and performing feature selection. In the later layers, the activation mean and variance reach higher levels, while sparsity decreases, implying that more neurons are activated to support information integration and complex reasoning required for final output generation. It is important to note that these observations may reflect characteristics specific to the evaluated datasets and models, and may not generalize across all settings.

\section{Analysis on Mixing Ratios}\label{app:Analysis on Mixing Ratios}
We further examine the effect of varying the ratio of long-context (ChatQA2) to short-context (UltraChat) SFT data, with results reported in Table~\ref{tab:mixing_ratios}. \textit{FFN Statistics} show that activation mean and variance rise initially and then decline as the proportion of long-context data increases, while sparsity decreases monotonically. This indicates that moderate amounts of long-context data enhance FFN expressiveness, whereas excessive use leads to over-regularization. \textit{Attention Entropy} increases during reasoning, suggesting improved exploration, but decreases during answering, reflecting stronger confidence. These trends are consistent with our main entropy analysis. \textit{Retrieval Score} peaks at extreme ratios (0:10 or 9:1) but deteriorates under intermediate mixtures, implying sensitivity to the dominant data type. Overall, these results highlight that FFN and MHA modules benefit from different optimal ratios, underscoring the importance of hybrid training.

\section{Analysis on Various Types of Factual Knowledge}
\label{app:various_factual}
We provide supplementary evaluations on various factual knowledge datasets. For the FFN module analysis, we extend the geographical dataset with 200 diverse questions sampled from Google’s \textit{Natural Questions }dataset, covering domains such as geography, culture, and history. As shown in Table~\ref{tab:nq_accuracy}, the results remain consistent with our main findings that long-context SFT data strengthens the FFN module.

We also extend the knowledge preference analysis by constructing three knowledge conflict datasets in the domains of \textit{Technology}, \textit{Celebrity}, and \textit{Sport}. Table~\ref{tab:knowledge_conflict} presents the results, which again confirm that the length of SFT training data introduces systematic biases in the model’s knowledge preferences.
\begin{table}[t]
\centering
\caption{Accuracy on Natural Questions dataset with 95\% confidence intervals.}
\label{tab:nq_accuracy}
\resizebox{\linewidth}{!}{%
\begin{tabular}{lcc}
\toprule
\textbf{Model} & \textbf{Accuracy (Mean $\pm$ Std)} & \textbf{95\% CI} \\
\midrule
ChatQA2 & 50.60 $\pm$ 0.82\%& $\pm$0.72\%\\
ChatQA2 + UltraChat FFN & 46.90 $\pm$ 0.96\%& $\pm$0.84\% \\
UltraChat & 40.70 $\pm$ 0.57\%& $\pm$0.50\%\\
UltraChat + ChatQA2 FFN & 45.70 $\pm$ 0.45\%& $\pm$0.39\%\\
\bottomrule
\end{tabular}%
}
\end{table}

\begin{table}[t]
\centering
\caption{Accuracy (\%) on knowledge conflict datasets across different training data mixing ratios~(UltraChat : ChatQA2).}
\label{tab:knowledge_conflict}
\resizebox{\linewidth}{!}{%
\begin{tabular}{lccc}
\toprule
\textbf{Ratio} & \textbf{Tech (\%)} & \textbf{Celebrity (\%)} & \textbf{Sport (\%)} \\
\midrule
10:0 & \textbf{59} & \textbf{65} &\textbf{77} \\
9:1  & 51 & 62 & 71 \\
8:2  & 44 & 49 & 61 \\
7:3  & 49 & 56 & 62 \\
6:4  & 50 & 60 & 64 \\
5:5  & 39 & 53 & 62 \\
4:6  & 42 & 49 & 61 \\
3:7  & 44 & 55 & 64 \\
2:8  & 39 & 51 & 58 \\
1:9  & 30 & 49 & 53 \\
0:10 & 34 & 44 & 48 \\
\bottomrule
\end{tabular}%
}
\end{table}

\begin{table*}[t]
\centering
\caption{Effect of mixing ratios~(UltraChat : ChatQA2) on FFN statistics, attention entropy, and retrieval score.}
\label{tab:mixing_ratios}
\resizebox{\linewidth}{!}{%
\begin{tabular}{ccccccc}
\toprule
\textbf{Ratio} & 
\textbf{$\Delta$ Mean (FFN) $\uparrow$} & 
\textbf{$\Delta$ Sparsity (FFN) $\downarrow$} & 
\textbf{$\Delta$ Variance (FFN) $\uparrow$} & 
\textbf{$\Delta$ Reasoning Entropy (Attn) $\uparrow$} & 
\textbf{$\Delta$ Answer Entropy (Attn) $\downarrow$} & 
\textbf{$\Delta$ Retrieval Score $\uparrow$} \\
\midrule
10:0 & 0.00 &  0.00 & 0.00 & 0.00 &   0.00 &  0.00 \\
9:1  & 0.29 & \textbf{-0.58} & 1.89 & 0.28 &  -7.80 &  0.34 \\
8:2  & \textbf{0.31} & -0.51 & \textbf{2.06} & 1.96 & -11.10 &  0.17 \\
7:3  & 0.30 & -0.44 & 1.91 & 2.40 & -14.20 &  0.09 \\
6:4  & 0.29 & -0.36 & 1.73 & 2.51 &  -9.47 &  0.08 \\
5:5  & 0.27 & -0.46 & 1.83 & 3.43 & -23.60 & -0.19 \\
4:6  & 0.24 & -0.41 & 1.83 & 1.54 & -38.10 & -0.00 \\
3:7  & 0.23 & -0.38 & 1.37 & 4.09 & -35.10 &  0.15 \\
2:8  & 0.20 & -0.23 & 1.26 & 4.65 & -41.00 &  0.17 \\
1:9  & 0.18 & -0.31 & 1.21 & 4.37 & -49.90 &  0.26 \\
0:10 & 0.15 & -0.48 & 0.88 & \textbf{6.86} & \textbf{-76.80} &  \textbf{0.43} \\
\bottomrule
\end{tabular}%
}
\end{table*}

\end{document}